\documentclass[letterpaper, 10 pt, journal, twoside]{IEEEtran}

\usepackage{graphics} 
\usepackage{float}
\usepackage[caption=false]{subfig}
\usepackage[final]{graphicx}
\usepackage{stackengine}

\usepackage{epsfig} 
\usepackage{mathptmx} 
\usepackage{times} 
\usepackage{amsmath,amssymb,bm} 
\usepackage{hyperref}
\usepackage[ruled,boxed,linesnumbered,vlined]{algorithm2e} 
\usepackage{booktabs,chemformula}

\usepackage{cleveref}

\newcommand{\figref}[1]{Fig.\ref{#1}}
\newcommand{\algoref}[1]{Algorithm \ref{#1}}

\newcommand{\sbf}{\mathbf{s}} 

\newcommand{\Pb}{\mathbf{P}}
\newcommand{\lmdb}{\bm\lambda}
\newcommand{\p}{\text{p}}

\title{Noticing Motion Patterns: 
A Temporal CNN with a Novel Convolution Operator 
for
Human Trajectory Prediction 
}

\author{Dapeng Zhao $^{1}$ and Jean Oh$^{1}$%
\thanks{Manuscript received: July 1, 2020; Revised November 12, 2020; Accepted December 7, 2020.}
\thanks{This paper was recommended for publication by Editor Tamim Asfour upon evaluation of the Associate Editor and Reviewers' comments.}
\thanks{
This work was supported in part by the Air Force Office of Scientific Research and U.S. Army Ground Vehicle Systems Center.} 
\thanks{$^{1}$Dapeng Zhao and Jean Oh are affiliated with the Robotics Institute, Carnegie Mellon University, Pittsburgh, Pennsylvania, USA 15213 
{\tt\small \{dapengz,hyaejino\}@andrew.cmu.edu}}
\thanks{Digital Object
 Identifier (DOI): see top of this page.}
}

\begin{document}

\maketitle

\IEEEpeerreviewmaketitle


\begin{abstract}
As more and more robots are envisioned to cooperate with humans sharing the same space, it is desired for robots to be able to predict others' trajectories to navigate in a safe and self-explanatory way. In this paper, we propose a Convolutional Neural Network-based approach to learn, detect, and extract patterns in sequential trajectory data, known here as Social Pattern Extraction Convolution (Social-PEC). A set of experiments carried out on the human trajectory prediction problem shows that our model performs comparably to the state of the art and outperforms in some cases. More importantly, the proposed approach unveils the obscurity in the previous use of a pooling layer, presenting a way to intuitively explain the decision-making process. 

\end{abstract}


\section{Introduction}\label{sec:intro}
\IEEEPARstart{W}{hether} 
an agent is co-working with other agents or is navigating in a crowd, it is critical for the agent to be able to understand and predict the motions of other agents in the same environment to ensure safe interaction and efficient performance. In this work, we particularly focus on the problem of pedestrian trajectory prediction in crowded environments. 
Predicting pedestrian behavior is challenging because pedestrians' future trajectories can be affected not only by the physical properties that we have well-established models to explain, such as energy or momentum, but also by the pedestrians' hidden objectives and subtle social norms in crowd interactions. 

\begin{figure}[t]\centering
    \subfloat[]{\includegraphics[trim=0 0 732 0,clip,scale=0.26]{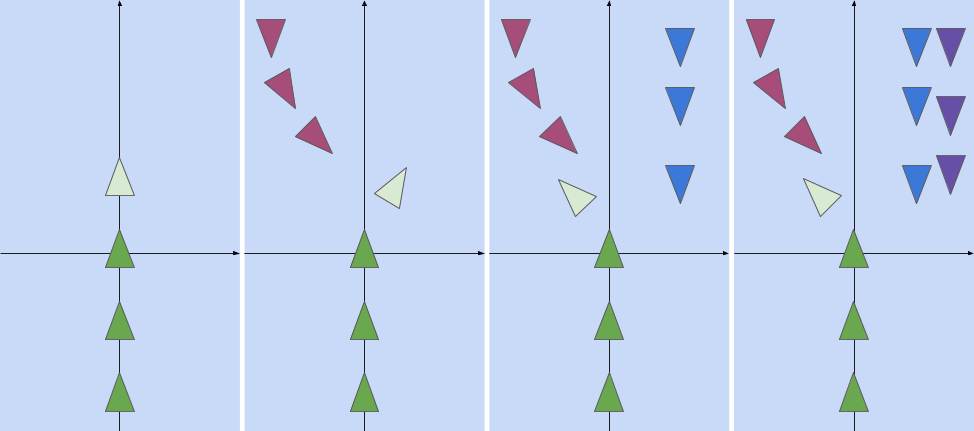}} 
    \subfloat[]{\includegraphics[trim=244 0 488 0,clip,scale=0.26]{figure/4scenario.png}}
    \subfloat[]{\includegraphics[trim=488 0 244 0,clip,scale=0.26]{figure/4scenario.png}}
    \subfloat[]{\includegraphics[trim=732 0 0 0,clip,scale=0.26]{figure/4scenario.png}} 
    \caption{Four sample scenarios of pedestrian interaction. The green triangles and the light green triangle represent the target pedestrian's history trajectories and expected one-step future location; triangles in other colors, the history trajectories of other pedestrians; and the sharp tips, the trajectory directions. In (a), no motion patterns from others are present; therefore, the target is expected to proceed linearly. In (b), the motion pattern of ``somebody approaching me from my front left'' is present, which impacts our target to walk to the right in order to avoid a possible collision. In (c) and (d), another motion pattern of ``somebody on my right is about to pass by me'' is present; therefore, our target pedestrian walks towards left. }
    \label{fig:4scenario}
    \end{figure}

The majority of existing work in pedestrian trajectory prediction generally follows the encoder-decoder model where the past trajectories are summarized to capture the context of the crowd movements. 
In this paper, we propose an approach, known here as Social Pattern Extraction Convolution (Social-PEC), where trajectories are eventually represented as a combination of \textit{motion patterns}.
Whereas existing approaches take the raw trajectories to encode the social context, the proposed idea on motion patterns is to generalize the observed trajectory as illustrated in Fig.~\ref{fig:4scenario}.
We design our model to understand and learn the various motion patterns of pedestrians in crowds, force the model to ``notice'' them during training and predict upon them. 
We build our sequence encoder using the idea of Temporal Convolutional Neural Networks (CNN)~\cite{lecun89cnn} and propose a new convolution operator (defined in \ref{sec:pattern})
that enables our model to actually detect, learn, and extract motion patterns from the observed trajectories. Using a different convolution operator is not a new idea: in~\cite{ghiasi19-generalizing}, the conventional correlation-based convolution operator has been modified to successfully achieve satisfying performance on the MNIST dataset, showcasing the applicability of the generalization of convolution operation in CNNs. 

Our model, Social-PEC, achieves comparable results with the state-of-the-art methods on public datasets in terms of standard evaluation metrics based on the displacement errors. Additionally, the use of motion patterns unveils the obscurity in social pooling, and makes the decision making process more transparent, intuitive, and explainable.


\section{Related Work}\label{sec:related}
Several technical approaches have been used to tackle the pedestrian trajectory prediction problem. In early works, algorithm designers tried to assert domain knowledge about social interactions in crowds to algorithms explicitly, e.g., Social Force~\cite{helbing95-socialForce} and Interactive Gaussian Process (IGP)~\cite{Trautman10-igp, Trautman13-multiGoal}. However, some researchers later suggested that hand-crafting models and rules have various limitations. These findings have led to those approaches that would allow the machines to learn directly from data~\cite{kitani12-af, alahi16-socialLstm, gupta18-socialGan}, resulting in significantly improved performances in general. We will mainly discuss the latter data-driven type of techniques in this section. 

For modeling sequential data, Recurrent Neural Network (RNN) and its variations such as Long Short-Term Memory network (LSTM) have been the popular choice, e.g.,  Social-LSTM~\cite{alahi16-socialLstm}; however, the benefit or necessity of using RNNs for pedestrian trajectory prediction is debatable. RNNs tend to gradually forget information from the past, hence the idea of LSTM was proposed as a remedy to selectively forget/remember. An LSTM is a reasonable choice in many problem domains where the sequences can be arbitrarily and extensively long, e.g., in the text-related tasks, a sequence could be a lengthy article where important contextual information can appear anywhere in the text. By contrast, in the pedestrian trajectory prediction problem, except for some rare extreme cases, the models do not need excessively long sequences as their inputs, as it is generally enough to observe how people have moved in the last 5 seconds, i.e., the observations from far too past are no longer meaningful to the current interaction, if not misleading at times. Moreover, RNNs/LSTMs have other drawbacks~\cite{bai18-cnnrnn} including vanishing/exploding gradients, unstable and expensive training, and inefficient parameters. Remarkable efforts were also made by Bai et al.~\cite{bai18-cnnrnn} and Becker et al.~\cite{becker18-comparison} to empirically evaluate RNNs for sequential data learning.  

After each sequence is modeled and encoded, information needs to be aggregated together, for which a pooling layer has been a popular choice. A pooling layer is widely used in CNNs for the image processing tasks, typically after spatial convolutional layers~\cite{krizhevsky12-alexnet}. In these CNNs, the vectors being pooled are quite literally the correlation between the kernels and the signals, and a pooling operation is to extract the strongest signal in a local region. However, the hidden space of RNNs/LSTMs is not well understood, and it is difficult, if not impossible, to understand its semantic meanings in a physical space. That said, it is yet to be justified to use a pooling layer after RNN/LSTM's modelling sequences for information aggregation. On this note, Mohamed et al.~\cite{mohamed20-stgcnn} reached a consensus with us. 

Another popular design choice is to use the graph representation. The combination of a graph with RNNs/LSTMs~\cite{vemula18-attention, kosaraju19-bigat}, and the combination of a graph with CNNs (Graph CNN)~\cite{mohamed20-stgcnn} have both been proposed. In these approaches, however, the size of a graph is generally dependent on the number of pedestrians in the scene, so it can face the scalability challenges as the number of pedestrians grows significantly in crowded scenes. 


An extensive and comprehensive survey article for human motion prediction is done by Rudenko et al.~\cite{rudenko2019-predSurvey}, where interested readers can find additional relevant works.

In contrast to the existing approaches, the proposed Social Pattern Extraction Convolution (Social-PEC) model avoids the issues pointed out above in this section. Firstly, it takes a fixed length of history trajectory as inputs and does not rely on RNNs to encode trajectory, thus avoids the RNN training issue. Secondly, the trajectory encoder is based on ``motion patterns'' that are intuitively reasonable and can be easily visualized, thus avoids the obscurity during ``social pooling'' which is to aggregate the encoded embedding of the neighboring pedestrians' trajectories.

\section{Trajectory Prediction with PEC} \label{sec:approach}

Our approach builds on an assumption that, when navigating in a crowd, humans react to an abstract representation of a scene, e.g., at the level of motion patterns that they have seen frequently in their past experiences. For instance, whether they are 2 or 3 people, whether they walk slightly faster or slower, whether they are a few centimeters to the left or right, as long as they approach from the same general direction at roughly the same distance with similar speeds, we probably will react very similarly as shown in~\figref{fig:4scenario}. Although trajectory data are mostly stored as sequences of location coordinates, humans do not react to precise location coordinates; instead, we react to general \emph{motion patterns}. 

In this paper, we define a ``pattern'' to be a segment of data. Specifically, a ``motion pattern'' refers to a short segment of trajectory that can be frequently seen in real trajectory data. A motion pattern is represented as a sequence of location coordinates, similar to a short trajectory representation. 

Logistically, our strategy is to only predict one-step future locations, and use the predicted locations as if they are the new observations to further predict. When predicting a one-step future, we predict for each pedestrian at a time. 
The pedestrian we predict for is referred to as a ``target pedestrian'',
while all others are ``context pedestrians'', and all trajectories are transformed from the world coordinates to the target pedestrian's egocentric coordinates where the target pedestrian is at the origin, facing the positive direction of x-axis.

In this section, we first clarify problem setup and notation; then introduce our model's key component, Pattern Extraction Convolution (PEC); next, we introduce the actual trajectory predictor; finally, we explain training and inference.

    \subsection{Representation and Notation}\label{sec:problem}

Suppose that there are $M$ pedestrians in a scene. Given all of their observed history trajectories in the world coordinates, the goal is to predict the future trajectories of all pedestrians. 

A trajectory is a series of states timestamped at a constant interval. State $\sbf$ is defined as a 2-dimensional location coordinates, s.t., $\sbf=(x,y)$. We note that the definition of a state can be extended to include additional information such as  orientation or personality in the future.

In this paper, we use a finite length $T_f$ of timesteps. The start and end time for history observations are $1$ and $T_h$, and that of future prediction are $T_h+1$ and $T_f$.

For pedestrian $m \in \{1, ..., M\}$, the trajectory of $m$ observed from timestep $1$ to timestep $T_h$, denoted by $\phi^m_{1:T_h}$, is composed of a sequence of $(x,y)$ coordinates as follows: 
\begin{align}
    \phi^m_{1:T_h} &= \{\sbf^m_t|t\in\{1,..,T_h\}\}
\end{align} where $\sbf^m_t= (x^m_t,y^m_t)$ is pedestrian $m$'s state at time $t$.

Let $\Phi$ denote the set of trajectories, s.t., $\Phi[m]=\phi^m, \forall m \in \{1,...,M\}$; the dimension of $\Phi$ is thus $(M,T_f,2)$.


    \subsection{Pattern Extraction Convolution (PEC)}\label{sec:pattern}

Pattern Extraction Convolution (PEC) is a mechanism that, in an intuitive sense, detects and recognizes patterns from data, while effectively projects trajectories from the x-y coordinate space to a new space that is defined in terms of similarities between patterns and the trajectory.

Let $\Pb$ denote the set of motion patterns, and let there be $N$ patterns in total. The $j$-th motion pattern, $\forall j\in\{1,...,N\}$, is defined as the following:
\begin{align}
    \Pb[j] &= \{\sbf^j_t|t\in\{1,..,L\}\}
\end{align}
where $L$ is the pattern length, $\sbf^j_t=(x^j_t,y^j_t)$ is the state. The dimension of $\Pb$ is thus $(N,L,2)$. 

\begin{figure}[t]\centering
    \subfloat[]{\includegraphics[width=1.665in]{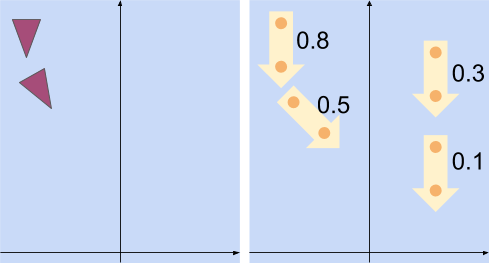}}\, 
    \subfloat[]{\includegraphics[width=1.665in]{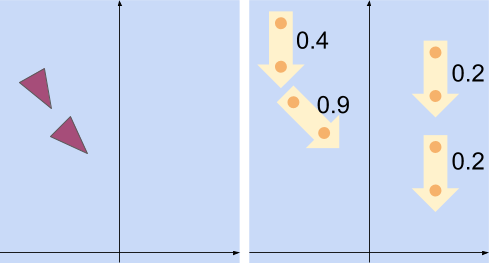}}\\ 
    \centering\subfloat[]{\centering\includegraphics[width=1.75in]{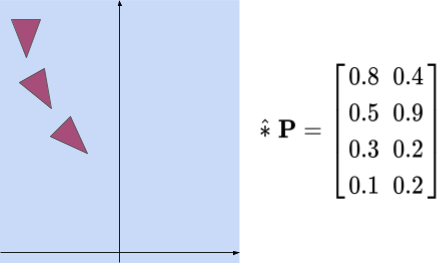}} 
    \caption{An example showing how raw trajectories can be projected to the motion pattern space. 
    Red triangles represent the input trajectory. The full trajectory $\phi$ is shown in (c) which is of shape $(3,2)$, for $T=3$. 
    Yellow arrows represent the set of motion patterns of shape $(4,2,2)$, for the number of patterns $N=4$, the length of pattern $L=2$, and $\sbf=(x,y)$ is 2-dimensional. 
    In (a), the similarities between the first segment of the trajectory and each motion pattern are found and marked next to the arrow; in (b), the second segment, similarly.
    The full operation on the trajectory level is shown in (c), where the resulting matrix is $\psi$ in the shape of (2,4) for $T_h-L+1=2$ and $N=4$. 
    The columns of this matrix $\phi$ are similarity scores from (a) and (b), respectively.%
    }
    \label{fig:conv}
\end{figure}

We define PEC as an encoder that translates raw trajectory $\phi$ to abstract trajectory $\psi$ in terms of motion patterns $\Pb$ as follows: 
\begin{equation}\psi = \text{PEC}(\phi;\Pb)\end{equation}
At each timestep $t$, for each motion pattern $j$, the PEC operation is defined as:
\begin{align}\label{eq:pec}
    \psi[t,j] &= \text{PEC}(\phi;\Pb)[t,j] \\
        &=\lmdb[j]\cdot\log( \sum_k^L \Big\|\phi[t-L+k,:]-\Pb[j,k,:]\Big\|_2 ) + \mathbf{b}[j] \nonumber
\end{align}
where $\lmdb$ is a scaling coefficient, and $\mathbf{b}$, biases. The dimension of the resulting encoded trajectory $\psi$ is $(T_h-L+1,N)$ where each entry of $\phi[t,j]$ indicates the similarity between the corresponding segment of the trajectory and pattern $P[j]$. This operation is further demonstrated in \figref{fig:conv}. 

Because the output $\psi$ of the PEC operation will later interact with an activation function such as \emph{tanh}, it is necessary to bring in the $\log$ function and the extra scaling coefficients $\lmdb$. Specifically, the $\log$ function helps re-range the values of the L2-distances from $[0,+\infty)$ back to $(-\infty,+\infty)$. Scaling coefficients $\lmdb$ helps scale the output properly to better interact with the non-linearity of the activation function. These two practices were not necessary for the conventional convolution operator
because the dot-product operation naturally ranges $(-\infty,+\infty)$ where the magnitude of a kernel matrix adds already an extra degree of freedom to help scale the response.

\begin{figure}[h]\centering
\includegraphics[width=1.8in]{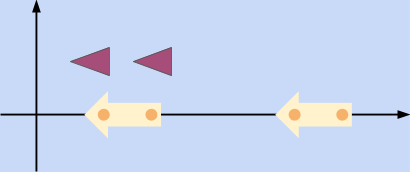} 
\caption{An example showing the incompetency of the conventional convolutional operator for trajectories. Suppose red trajectory $\phi$ is \{(10,1),(20,1)\}, and two patterns $\p_0$ and $\p_1$ are \{(10,0),(20,0)\} and \{(50,0),(60,0)\}. Ideally, the magnitude of the convolution response (or output) is determined by the similarity only, i.e., the larger the more similar, but if CONV is used the magnitude of response would also be influenced by the magnitude of the input signals or kernels. In the example, the similarity between $\phi$ and $\p_0$ should be larger than $\p_1$, but $\text{CONV}(\phi,\p_0)=500<1700=\text{CONV}(\phi,\p_1)$.}
\label{fig:conv_compare}
\end{figure}

It is worth noting that the PEC operator defined in~\eqref{eq:pec} is different from the conventional convolution (CONV) operator~\cite{ghiasi19-generalizing,paszke19-pytorch} which is commonly used in CNNs for image-related tasks. The main differences are (1) PEC is based on the L2-difference to measure the physical distance whereas CONV is based on dot-product; (2) the conventional operator ignores the physical meaning of the channels and simply sums up the outputs from different channels. The necessity for using the PEC operator for trajectory encoding is further illustrated in~\figref{fig:conv_compare}.
%

The set of motion pattern, $\Pb$. is learned from data, trained using the prediction loss through back propagation. More implementation details and illustration are presented in Sections~\ref{sec:implementation} and~\ref{sec:show_steps}.

    \subsection{Human Trajectory Predictor}\label{sec:model}
    
    \begin{algorithm} \label{algo:workflow}
    \DontPrintSemicolon
    $\Phi \gets \{\phi^m|m\in[1:M]\}$\;
    \SetKwFunction{TrajPredictor}{TrajPredictor}
    \SetKwFunction{Trans}{Convert}
    \SetKwFunction{TransBack}{ConvertBack}
    \SetKwFunction{LocPredictor}{LocPredictor}
    \SetKwProg{Fn}{Function}{:}{}
    \Fn{\TrajPredictor{$\Phi$}}{
    \For{$t=T_h+1,..,T_f$}{
        \For{$m=1,..,M$}{
            $\Phi'\gets\Trans(\Phi[:,t-T_h:t], m)$\;
            $\mu',\Sigma'\gets\LocPredictor(\Phi'[m],\Phi'[-m])$\;
            $\sbf'\sim\mathcal{N}(\mu',\Sigma')$\;
            $\sbf^m\gets\TransBack(\sbf') $\;
            $\Phi[m,t]\gets\sbf^m$\;        
            }
    }
    \KwRet $\Phi[:,T_h+1:T_f]$\;
    }
    \caption{Trajectory Predictor}
    \end{algorithm}
    
    Our strategy is to predict one-step future locations first, and then use the predicted locations as if they are new observations to further predict. At each timestep, every pedestrian is treated as target pedestrian by turns. 
    
    For each target pedestrian $m \in \{1, ..., M\}$, the location coordinates of all trajectories are first transformed from the world coordinates to the new coordinates where pedestrian $m$ is at the origin, oriented towards the positive direction of the x-axis, as in the \texttt{Convert} function in~\algoref{algo:workflow}. Next, Location Predictor takes all of these trajectories as inputs, and outputs $m$'s next location, which is modeled as bivariate Gaussian distribution~\cite{alahi16-socialLstm}. Finally, this location is converted back to the world coordinates and appended to $\Phi$ for location prediction for the next timestep.
    
    \begin{figure}\centering
        \includegraphics[width=3.3in]{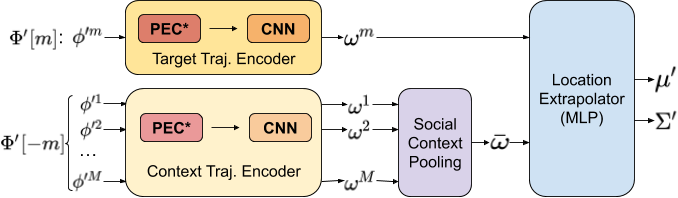}
        \caption{Location Predictor that predicts one-step location for pedestrian $m$. $\Phi'$ is the set of trajectories already transformed to target pedestrian's coordinates. $\omega$ is the trajectory embedding, explained in Equation~\eqref{eq:encoder}. }
        \label{fig:loc_predictor}
    \end{figure}

    Given target pedestrian $m$, let $-m$ denote the rest of the pedestrians except $m$, referred to as Context.
    First, $\Phi'[m]$ and $\Phi'[-m]$ are encoded respectively by two different encoder networks. The reason for having two different networks is that motion pattern set and context trajectories are expected to be different for Context and Target. Trajectory Encoder works on one trajectory at a time. The Target trajectory embedding $\omega$ can be written as follows:
    \begin{align}\label{eq:encoder}
        \omega &= \text{CNN}(\sigma(\psi)) \nonumber\\
        &= \text{CNN}(\sigma(\text{PEC}(\phi;\Pb)))
    \end{align}
    where $\phi$ is the raw trajectory in location coordinates space;
    $\psi$, the trajectory encoded by PEC;
    and
    $\sigma$, the activation function. 
    By applying CNN to the encoded trajectory $\psi$, combinations of basic motion patterns are further extracted by the CNN on higher levels as more sophisticated patterns.
    
    Now that the model is ready to predict future trajectory for target pedestrian, observations of all other pedestrians should be aggregated to provide the social context. Context Pooling layer is applied to compute context trajectory $\bar\omega$ as follows:
    \begin{equation}\label{eq:pool}
        \bar\omega[t,k] = \max_{m\in-m} (\omega^m[t,k])
    \end{equation}
    where $\bar\omega$ has the same shape as the target's encoded trajectory $\omega^m$, $t$ and $k$ are time and pattern indices. 
    
    To make predictions, both encoded target trajectory $\omega^m$ and social context $\bar\omega$ are fed into  Multilayer Perceptron (MLP) in the end:
    \begin{equation}\label{eq:mlp}
        x,y,a,b,c = \text{MLP}(\omega^m,\bar\omega).
    \end{equation}
    The $\mu'$ and $\Sigma'$ of location's Gaussian distribution are then constructed from the raw MLP outputs:
    \begin{equation}\label{eq:gaussian}
        \begin{matrix}
            \mu' = [x,y]\\[0.3em]
            \sigma'_{xx}, \sigma'_{yy} = \exp(a), \exp(b)\\[0.3em]
            \sigma'_{xy} = \sigma'_{xx}*\sigma'_{yy}*\tanh(c)\\[0.4em]
            \Sigma' = \begin{bmatrix}\sigma'_{xx}&\sigma'_{xy}\\\sigma'_{xy}&\sigma'_{yy}\end{bmatrix}
        \end{matrix}
    \end{equation}
    where superscript $'$ indicates that variable is in the target-centered coordinates instead of the world coordinates. Construction in \eqref{eq:gaussian} is necessary because the value range of raw MLP outputs, $[-\infty,\infty]$, does not satisfy the constraints of the covariance matrix. 



    \subsection{Training and Inferencing}

During training, the future prediction length is set to be 1, s.t., $T_f=T_h+1$,
because all parameters that need to be trained are all within Location Predictor. 
    
    Parameters are learned by minimizing the following negative log-likelihood loss:
    \begin{equation}
        L = - \sum_{m=1}^M\log(\,P(\Phi'[m,T_h+1]\,\big\vert\,\mu'^m,\Sigma'^m)\,)
    \end{equation}

    During inferencing, the future prediction can have an arbitrary length, and the prediction output is $\Phi[:,T_h+1:T_f]$ as stated in~\algoref{algo:workflow}.

\section{Evaluation and Discussion}\label{sec:eval}
\newcommand{\ii}{\textit}
\newcommand{\bb}{\textbf}
    
\begin{table*}\centering
    \caption{ADE/FDE in meters for different methods, the lower the better. Linear Extrapolation is included as baseline and its output is deterministic, whereas all other models predict 20 trajectories at once, and only the best one is counted for evaluation.}
    \resizebox{0.85\textwidth}{!}{
    \begin{tabular}{c||c|c|c|c|c||c}
        Model & ETH & Hotel & Univ. & Zara1 & Zara2 & Ave.\\\hline\hline
        Linear & 1.33 / 2.94 & 0.39 / 0.72 & 0.82 / 1.59 & 0.62 / 1.21 & 0.77 / 1.48 & 0.79 / 1.59 \\\hline
        S-LSTM\cite{alahi16-socialLstm} & 1.09 / 2.35 & 0.79 / 1.76 & 0.67 / 1.40 & 0.47 / 1.00 & 0.56 / 1.17 & 0.72 / 1.54 \\\hline
        SGAN(20VP20)\cite{gupta18-socialGan} & 0.87 / 1.62 & 0.67 / 1.37 & 0.76 / 1.52 & 0.35 / 0.68 & 0.42 / 0.84 & 0.61 / 1.21 \\\hline
        STSGN\cite{zhang19-stsgn} & 0.75 / 1.63 & 0.63 / 1.01 & 0.48 / 1.08 & \bb{0.30} / 0.65 & \bb{0.26} / 0.57 & 0.48 / 0.99 \\\hline
        S-BiGAT\cite{kosaraju19-bigat} & 0.69 / 1.29 & 0.49 / 1.01 & 0.55 / 1.32 & \bb{0.30} / 0.62 & 0.36 / 0.75 & 0.48 / 1.00 \\\hline
        S-STGCNN\cite{mohamed20-stgcnn} & 0.64 / \bb{1.11} & 0.49 / 0.85 & \bb{0.44} / \bb{0.79} & 0.34 / \bb{0.53} & 0.30 / \bb{0.48} & 0.44 / \bb{0.75} \\\hline
        \hline
        \bb{Social-PEC} & \bb{0.61} / \bb{1.11} & \bb{0.31} / \bb{0.52} & 0.47 / 0.82 & 0.43 / 0.77 & 0.35 / 0.60 & \bb{0.43} / 0.76 \\\hline
    \end{tabular} 
    }
    \label{table:afde}
\end{table*}

\subsection{Datasets and Metrics}
Our model is evaluated on two datasets:~\cite{pellegrini09-eth} and~\cite{lerner07-ucy}. They contain 5 crowd sets in different scenes with a total number of 1,536 pedestrians exhibiting complex interactions such as walking together, groups crossing each other, joint collision avoidance and nonlinear trajectories. 

As for metrics, like~\cite{alahi16-socialLstm, gupta18-socialGan, zhang19-stsgn, kosaraju19-bigat, mohamed20-stgcnn}, we use Average/Final Displacement Error (ADE/FDE)~\cite{mohamed20-stgcnn}, which have been conventionally used for this problem. 

In order to make full use of the data for evaluation and also to evaluate how well models generalize to unseen datasets, we use the leave-one-out approach where a model is trained and validated on 4 datasets and tested on the remaining set. To ensure a fair comparison, we use identical dataset step and train/validation/test split which are also used in~S-SLSTM\cite{alahi16-socialLstm}, S-GAN\cite{gupta18-socialGan} and S-STGCNN\cite{mohamed20-stgcnn}.

The data used in our work are annotated every 0.4 seconds. Observation length is set to be 8 timesteps (3.2 sec) and future prediction length is set to be 12 timesteps (4.8 sec).

\subsection{Implementation Details}\label{sec:implementation}
As shown in \figref{fig:loc_predictor}, our model mainly contains 3 modules, Context Trajectory Encoder, Target Trajectory Encoder, and Location Extrapolator. 

Context Trajectory Encoder consists of a Pattern Extraction Convolution (PEC) layer and a conventional convolution (CONV) layer, each followed by activation function \emph{tanh}, with a max pooling layer in between. For the PEC and CONV, the number of kernels are 100 and 160, the kernel lengths are 2 and 2; the pooling stride is 2. If the number of input channels is 2 (x-y coordinates) and the data temporal length is 8 (length of observations), s.t. an input is in the shape of (2,8), the resulting output's shape will be (160,3). 

Target Trajectory Encoder is very similar to Context Trajectory Encoder, except that the number of kernels are only 50 and 80, because there are much less varieties among target trajectories because the irrelevant variance have been removed by transforming the coordinate system with respect to target trajectory in \texttt{Convert} of \algoref{algo:workflow}.

Location Extrapolator are 4 fully-connected layers, followed by leaky Re-LU activation. The width of the layers are 300, 120, 80, and 5. 

For training, the batch size is 64, and it trains for 150 epochs using the Adam Optimizer~\cite{kingma14-adam} with the learning rate set as 0.001. The model is trained on GeForce RTX 2080 Ti.

\subsection{Quantitative Results}
As shown in Table 1, our model outperforms almost all models and performs comparably well with the current state-of-the-art, Social-STGCNN~\cite{mohamed20-stgcnn}. Other works are mostly RNN-based, while ours and Social-STGCNN are CNN-based. Our results suggest that CNNs might be a better option for modeling some types of sequences, e.g., pedestrian trajectories; more in-depth discussions related to the comparison between CNN and RNN can also be found in~\cite{becker18-comparison} and~\cite{bai18-cnnrnn}. 


\subsection{Qualitative Analysis}
\begin{figure}[t]\centering
    \newcommand{\widthR}{1.66in}
    \stackunder[0pt]{\includegraphics[trim=.56in .32in .63in .58in, clip,width=\widthR]{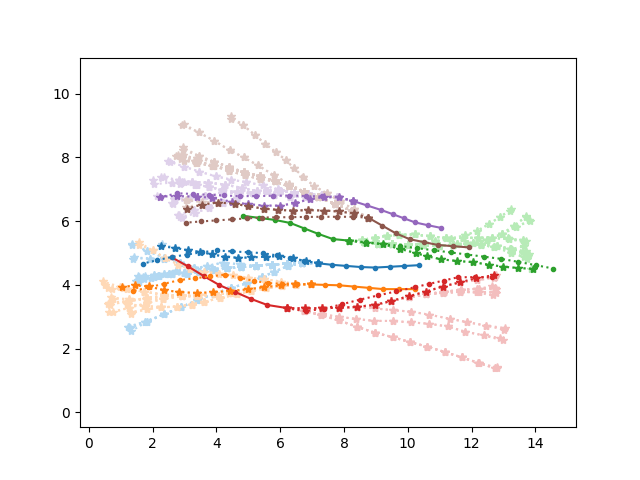}}{\footnotesize{(a)}}
    \stackunder[0pt]{\includegraphics[trim=.95in 0.52in 1.05in .95in, clip,width=\widthR]{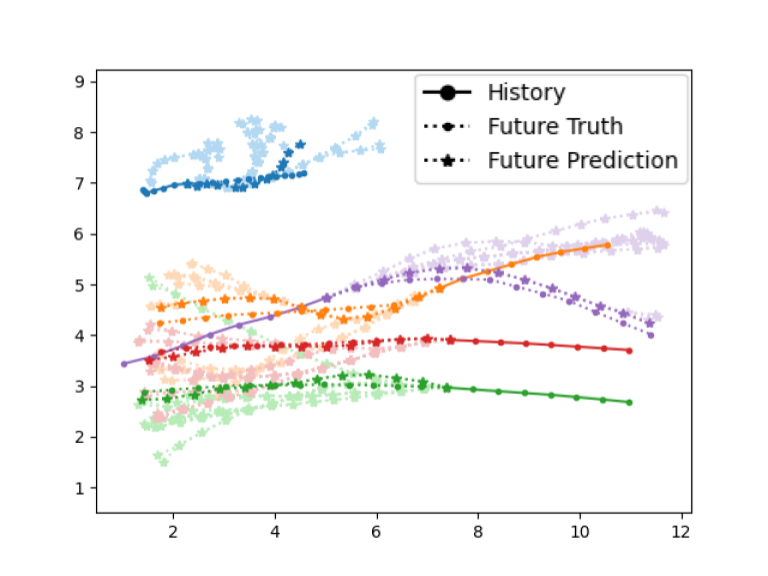}}{\footnotesize{(b)}}
    \stackunder[0pt]{\includegraphics[trim=.56in .32in .63in .58in, clip,width=\widthR]{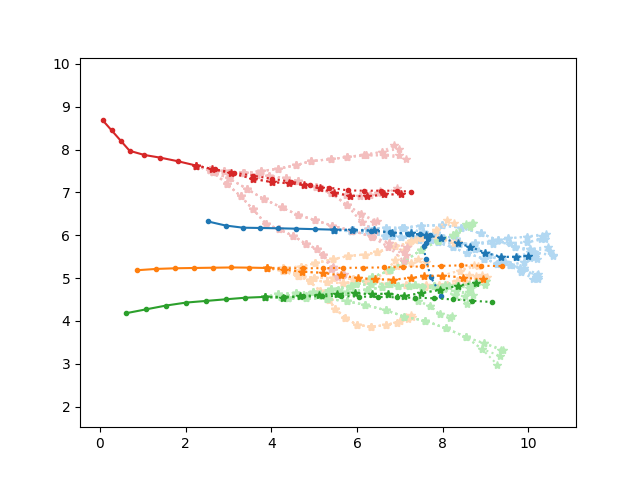}}{\footnotesize{(c)}}
    \stackunder[0pt]{\includegraphics[trim=.56in .32in .63in .58in, clip,width=\widthR]{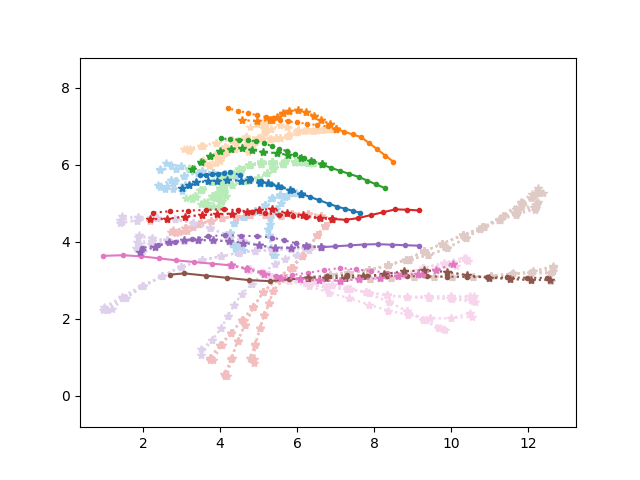}}{\footnotesize{(d)}}
    \stackunder[0pt]{\includegraphics[trim=.56in .32in .63in .58in, clip,width=\widthR]{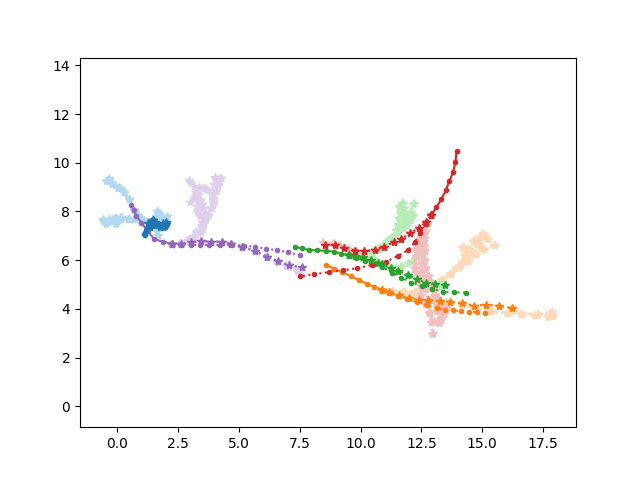}}{\footnotesize{(e)}}
    \stackunder[0pt]{\includegraphics[trim=.56in .32in .63in .58in, clip,width=\widthR]{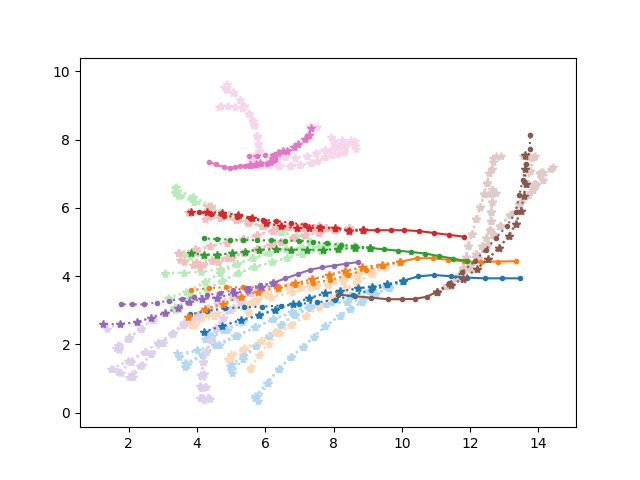}}{\footnotesize{(f)}}
    \caption{Some samples of our model's prediction. 
    Our model predicts 20 possible future trajectories for each pedestrian, of which the one with the smallest ADE is highlighted. This figure is best viewed in color. }
    \label{fig:results}
\end{figure}

Some sample results are shown in \figref{fig:results}. 
Typically, linear trajectories are trivial to predict; however, our results seem to support that the proposed model also performs well for some non-linear trajectories, especially in more crowded scenes, e.g., red in (a), purple and orange in (b), and brown in (f). 
The success here might indicate that the proposed model is able to make use of social context effectively to make more accurate and more reasonable predictions.

Some of the predictions deviate from the true future trajectories significantly, e.g., brown in (a), red in (e), and pink in (f). These predictions, however, still appear reasonable, that is, based on the observed trajectories, the prediction may appear arguably more reasonable than the true future. Such ``mispredictions'' are inevitable to some extent as some of the observed history trajectories might not carry enough information to allow anyone to accurately predict their future. 

Sometimes, the proposed is able to recognize the groups in crowd and predict accordingly, although we did not explicitly design the model to incorporate social group awareness~\cite{yao19-group}. In (f), the history observations of orange and blue are highly similar, thus their future trajectories are predicted to be very similar too, even the way how they deviate from ground truth are also similar. Comparatively, also in (f), the history observations of red and green did not demonstrate enough similarity, thus in the model's prediction they do not walk together any more. 

In some cases, the model successfully shows appropriate precaution for collision avoidance. In (e), the interaction between red and green is notable. Red took a different path that is farther away from green and both were predicted to move slower than ground truth, a plausible explanation is that the proposed model was trying to avoid the two pedestrians colliding with each other. In the same scene, the orange is predicted to move faster than ground truth, because its front space appears to be clear enough to allow faster speed.

\subsection{Motion Pattern Illustration}\label{sec:show_steps}

\begin{figure}[t]\centering
    \newcommand{\widthR}{1.66in}
    \stackunder[0pt]{\includegraphics[trim=.76in .65in .76in .96in,clip,width=\widthR]{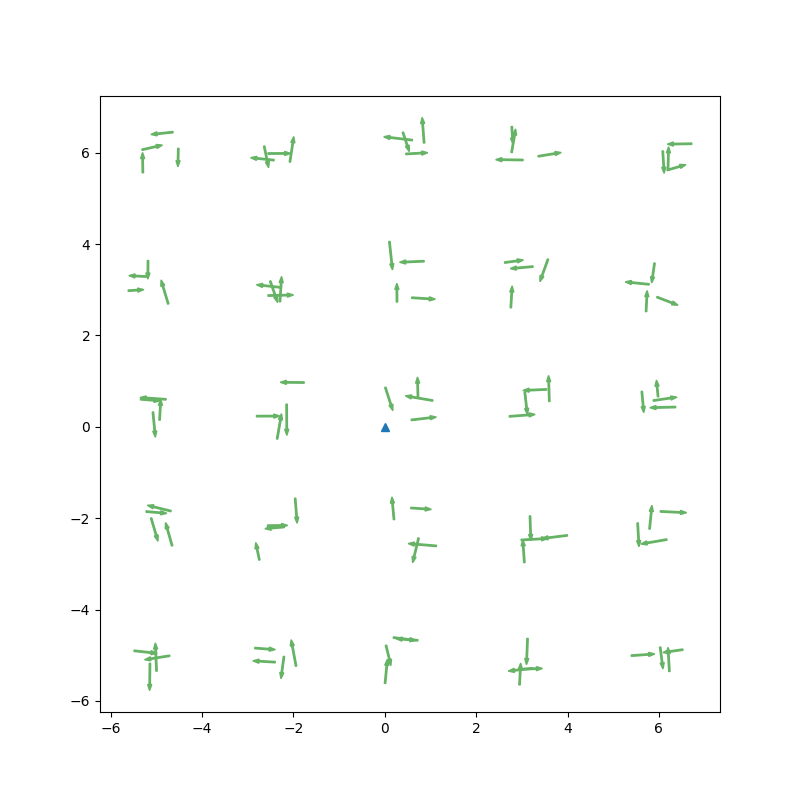}}{\footnotesize{(a)}}
    \stackunder[0pt]{\includegraphics[trim=.76in .65in .76in .96in,clip,width=\widthR]{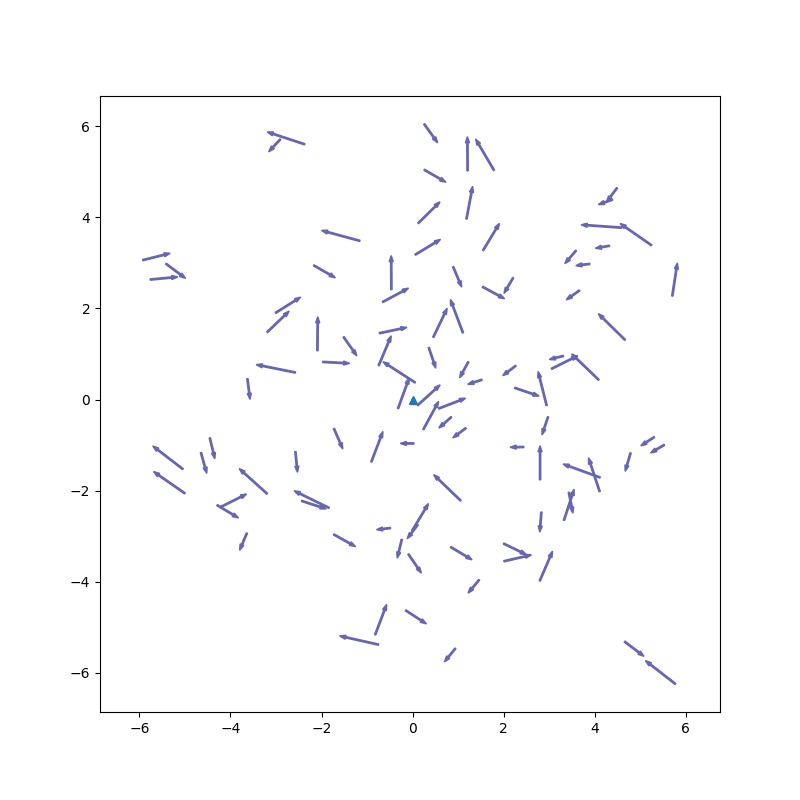}}{\footnotesize{(b)}}
    \stackunder[0pt]{\includegraphics[trim=.76in .65in .76in .96in,clip,width=\widthR]{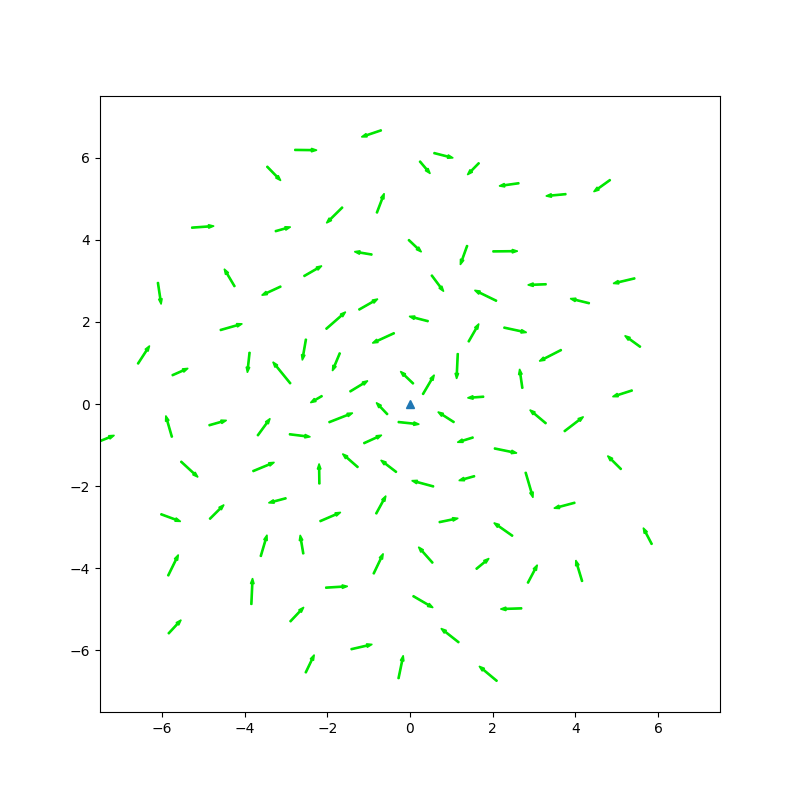}}{\footnotesize{(c)}}
    \stackunder[0pt]{\includegraphics[trim=.76in .65in .76in .96in,clip,width=\widthR]{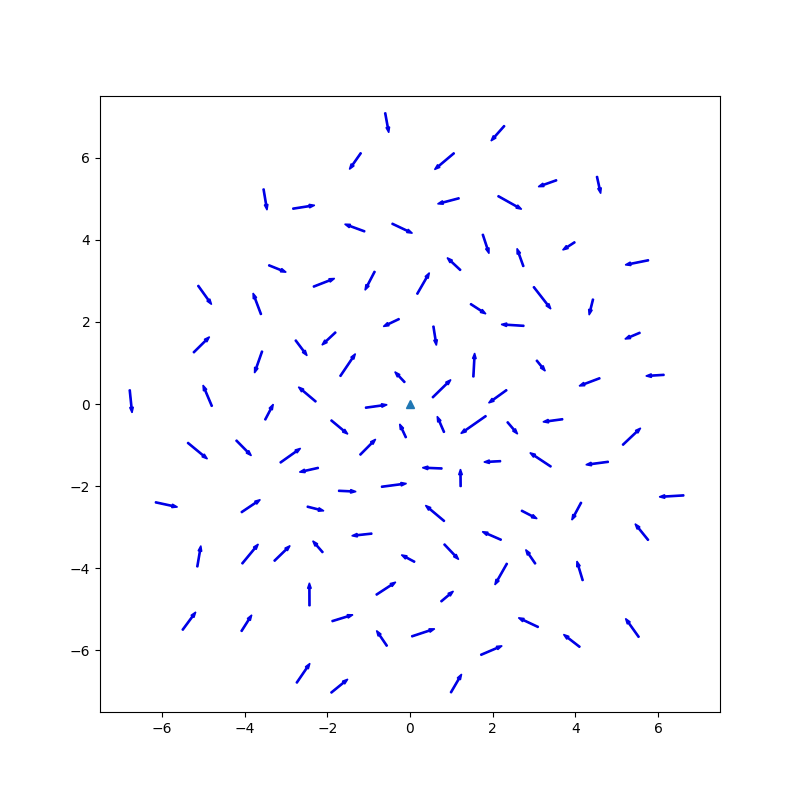}}{\footnotesize{(d)}}
    \caption{Two sets of initialization and learned motion patterns for Context Trajectory Encoder. (a) and (b) are initialization, while (c) and (d) are their corresponding learned patterns after training. The number of patterns is 100; length of patterns is 2. The coordinates are with respect to the target pedestrian marked by a blue triangle at (0,0). }
    \label{fig:init}
\end{figure}

\begin{figure*}[h]\centering
    \includegraphics[width=6in]{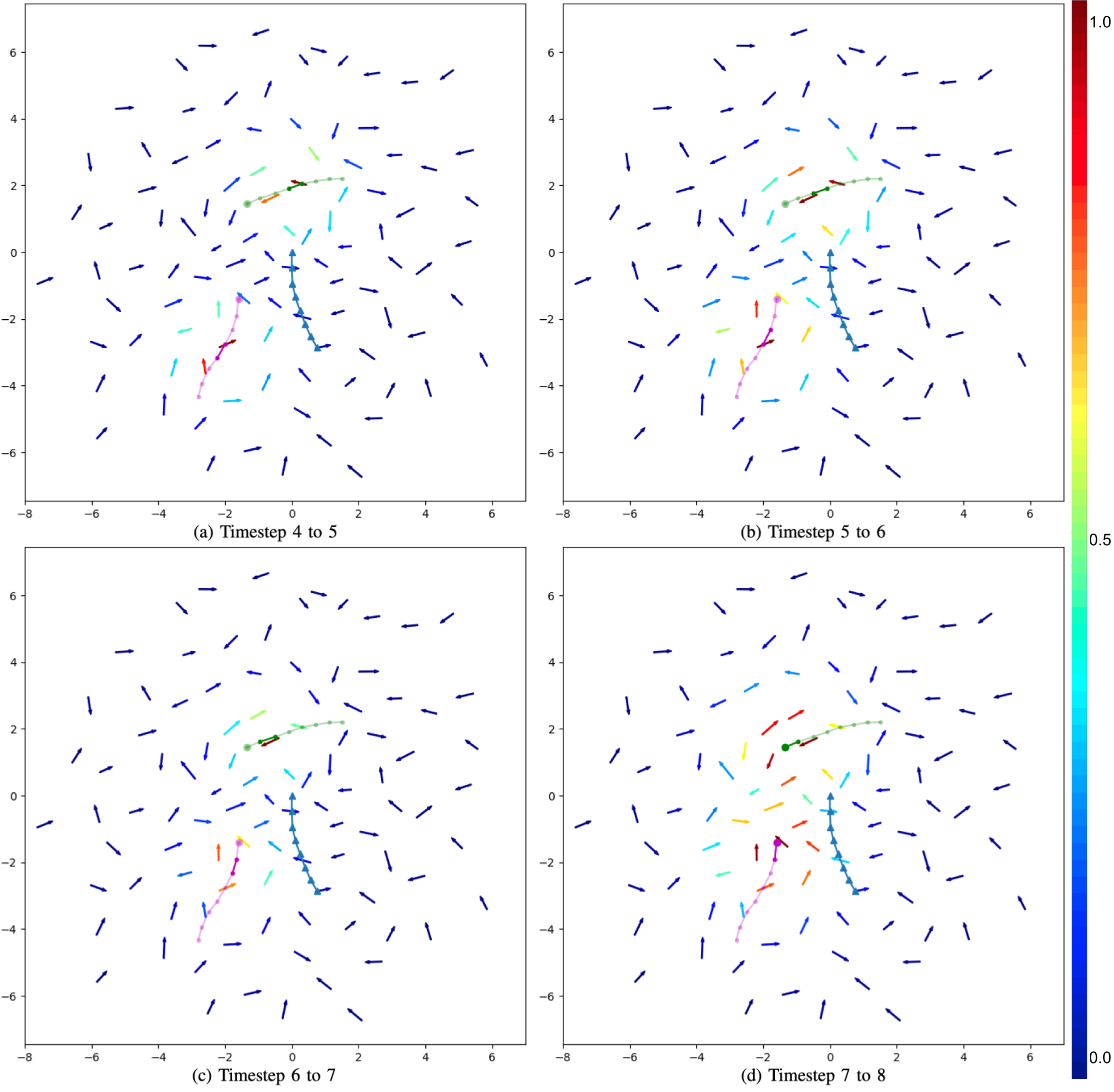}
    \caption{``Noticing motion patterns'' when presented trajectory data. The arrows are from \figref{fig:init}(c). All trajectories are observed history. Dots are context trajectories and larger round dots indicate the last-timestep locations. Blue triangles are the target trajectory which is only plotted but not matched to the learned patterns because patterns here are for context trajectories only. Trajectories are 8-timestep long, but the first 3 timesteps are neglected. Each subgraph demonstrates the matching of the particular trajectory segment, e.g., (d) shows the last segment, (c) shows the 2nd last, etc. By Pattern Extraction Convolution (PEC) layer, the segment is matched to similar patterns, causing these ``noticed patterns'' to be shown in heated color. }
    \label{fig:show_steps}
\end{figure*}

The weights for the PEC layer in Trajectory Encoder are motion patterns. \figref{fig:init} presents two sets of patterns trained from two very different initialization, which demonstrates that the training is robust enough to allow different initialization to converge to similar patterns. It is noted that the learned patterns are not identical though similar; however, no significant performance difference has been observed in terms of ADE/FDE.

\figref{fig:show_steps} illustrates the core idea of ``Noticing Motion Patterns'' with an example scenario, where observed history trajectory data are matched to learned motion patterns at each timestep. This is done by the first layers of Trajectory Encoders, Pattern Extraction Convolution (PEC) layers. 

 Under the scheme of ``motion pattern'', social pooling finally is no longer obscure. We are not pooling in any unknown latent space, but pooling in well-understood pattern space. The physical meaning of each entry in tensor $\omega$ is the similarity indicating how much of that particular motion pattern is present in the current scene. The larger the entry value is, the more similar the raw trajectory is to the motion pattern. Thus, stronger motion patterns should have a bigger impact on the target pedestrian's decision making. By only considering the prominent presence for each motion pattern, the model can already be well-informed about its social context, of which \figref{fig:4scenario}(c)(d) are good examples. 

\section{Conclusion}\label{sec:conclusion}

In this paper, we propose a CNN-based model for human pedestrian trajectory prediction with the idea of motion patterns. 
The main contributions of this work include:
\begin{itemize}
    \item we present Pattern Extraction Convolution (PEC) as an intuitive and explainable mechanism to learn, detect, and extract patterns from data, which is used to encode trajectories in this work;
    \item we further apply PEC to human trajectory prediction problem as the model of Social-PEC, and achieve comparable performance with the current state-of-the-art;
    \item the use of PEC avoids the obscurity in information aggregation (pooling layer) that was present in the previous work; and
    \item this study further challenges the community to re-examine the use of RNN in sequential data learning tasks.
\end{itemize}

In the future, for the problem of human motion prediction, we will try to incorporate physical environmental constraints into our model. 
Since the proposed idea on the PEC encoder is more general than the scope of trajectory prediction, we plan to apply PEC to other problem domains and further explore its potential.


\section*{Acknowledgement}
This work is in part supported by the U.S. Air Force Office of Scientific Research under award number FA2386-17-1-4660 and U.S. Army Ground Vehicle Systems Center.

\bibliographystyle{IEEEtran}
\bibliography{references}

\begin{thebibliography}{10}
\providecommand{\url}[1]{#1}
\csname url@samestyle\endcsname
\providecommand{\newblock}{\relax}
\providecommand{\bibinfo}[2]{#2}
\providecommand{\BIBentrySTDinterwordspacing}{\spaceskip=0pt\relax}
\providecommand{\BIBentryALTinterwordstretchfactor}{4}
\providecommand{\BIBentryALTinterwordspacing}{\spaceskip=\fontdimen2\font plus
\BIBentryALTinterwordstretchfactor\fontdimen3\font minus
  \fontdimen4\font\relax}
\providecommand{\BIBforeignlanguage}[2]{{%
\expandafter\ifx\csname l@#1\endcsname\relax
\typeout{** WARNING: IEEEtran.bst: No hyphenation pattern has been}%
\typeout{** loaded for the language `#1'. Using the pattern for}%
\typeout{** the default language instead.}%
\else
\language=\csname l@#1\endcsname
\fi
#2}}
\providecommand{\BIBdecl}{\relax}
\BIBdecl

\bibitem{lecun89cnn}
Y.~LeCun, B.~Boser, J.~S. Denker, D.~Henderson, R.~E. Howard, W.~Hubbard, and
  L.~D. Jackel, ``Backpropagation applied to handwritten zip code
  recognition,'' \emph{Neural computation}, vol.~1, no.~4, pp. 541--551, 1989.

\bibitem{ghiasi19-generalizing}
K.~Ghiasi-Shirazi, ``Generalizing the convolution operator in convolutional
  neural networks,'' \emph{Neural Processing Letters}, vol.~50, no.~3, pp.
  2627--2646, 2019.

\bibitem{helbing95-socialForce}
D.~Helbing and P.~Molnar, ``Social force model for pedestrian dynamics,''
  \emph{Physical review E}, vol.~51, no.~5, pp. 4282--4286, 1995.

\bibitem{Trautman10-igp}
P.~{Trautman} and A.~{Krause}, ``Unfreezing the robot: Navigation in dense,
  interacting crowds,'' in \emph{2010 IEEE/RSJ International Conference on
  Intelligent Robots and Systems}, 2010, pp. 797--803.

\bibitem{Trautman13-multiGoal}
P.~{Trautman}, J.~{Ma}, R.~M. {Murray}, and A.~{Krause}, ``Robot navigation in
  dense human crowds: the case for cooperation,'' in \emph{2013 IEEE
  International Conference on Robotics and Automation}, 2013, pp. 2153--2160.

\bibitem{kitani12-af}
K.~M. Kitani, B.~D. Ziebart, J.~A. Bagnell, and M.~Hebert, ``Activity
  forecasting,'' in \emph{European Conference on Computer Vision}.\hskip 1em
  plus 0.5em minus 0.4em\relax Springer, 2012, pp. 201--214.

\bibitem{alahi16-socialLstm}
A.~Alahi, K.~Goel, V.~Ramanathan, A.~Robicquet, L.~Fei-Fei, and S.~Savarese,
  ``Social lstm: Human trajectory prediction in crowded spaces,'' in
  \emph{Proceedings of the IEEE conference on computer vision and pattern
  recognition}, 2016, pp. 961--971.

\bibitem{gupta18-socialGan}
A.~Gupta, J.~Johnson, L.~Fei-Fei, S.~Savarese, and A.~Alahi, ``Social gan:
  Socially acceptable trajectories with generative adversarial networks,'' in
  \emph{Proceedings of the IEEE Conference on Computer Vision and Pattern
  Recognition}, 2018, pp. 2255--2264.

\bibitem{bai18-cnnrnn}
S.~Bai, J.~Z. Kolter, and V.~Koltun, ``An empirical evaluation of generic
  convolutional and recurrent networks for sequence modeling,'' \emph{arXiv
  preprint arXiv:1803.01271}, 2018.

\bibitem{becker18-comparison}
S.~Becker, R.~Hug, W.~H{\"u}bner, and M.~Arens, ``An evaluation of trajectory
  prediction approaches and notes on the trajnet benchmark,'' \emph{arXiv
  preprint arXiv:1805.07663}, 2018.

\bibitem{krizhevsky12-alexnet}
A.~Krizhevsky, I.~Sutskever, and G.~E. Hinton, ``Imagenet classification with
  deep convolutional neural networks,'' in \emph{Advances in neural information
  processing systems}, 2012, pp. 1097--1105.

\bibitem{mohamed20-stgcnn}
A.~Mohamed, K.~Qian, M.~Elhoseiny, and C.~Claudel, ``Social-stgcnn: A social
  spatio-temporal graph convolutional neural network for human trajectory
  prediction,'' in \emph{Proceedings of the IEEE/CVF Conference on Computer
  Vision and Pattern Recognition}, 2020, pp. 14\,424--14\,432.

\bibitem{vemula18-attention}
A.~{Vemula}, K.~{Muelling}, and J.~{Oh}, ``Social attention: Modeling attention
  in human crowds,'' in \emph{2018 IEEE International Conference on Robotics
  and Automation (ICRA)}, May 2018, pp. 4601--4607.

\bibitem{kosaraju19-bigat}
V.~Kosaraju, A.~Sadeghian, R.~Mart{\'\i}n-Mart{\'\i}n, I.~Reid, H.~Rezatofighi,
  and S.~Savarese, ``Social-bigat: Multimodal trajectory forecasting using
  bicycle-gan and graph attention networks,'' in \emph{Advances in Neural
  Information Processing Systems}, 2019, pp. 137--146.

\bibitem{rudenko2019-predSurvey}
A.~Rudenko, L.~Palmieri, M.~Herman, K.~M. Kitani, D.~M. Gavrila, and K.~O.
  Arras, ``Human motion trajectory prediction: A survey,'' \emph{The
  International Journal of Robotics Research}, vol.~39, no.~8, pp. 895--935,
  2020.

\bibitem{paszke19-pytorch}
A.~Paszke, S.~Gross, F.~Massa, A.~Lerer, J.~Bradbury, G.~Chanan, T.~Killeen,
  Z.~Lin, N.~Gimelshein, L.~Antiga \emph{et~al.}, ``Pytorch: An imperative
  style, high-performance deep learning library,'' in \emph{Advances in neural
  information processing systems}, 2019, pp. 8026--8037.

\bibitem{zhang19-stsgn}
L.~Zhang, Q.~She, and P.~Guo, ``Stochastic trajectory prediction with social
  graph network,'' \emph{arXiv preprint arXiv:1907.10233}, 2019.

\bibitem{pellegrini09-eth}
S.~Pellegrini, A.~Ess, K.~Schindler, and L.~Van~Gool, ``You'll never walk
  alone: Modeling social behavior for multi-target tracking,'' in \emph{2009
  IEEE 12th International Conference on Computer Vision}.\hskip 1em plus 0.5em
  minus 0.4em\relax IEEE, 2009, pp. 261--268.

\bibitem{lerner07-ucy}
A.~Lerner, Y.~Chrysanthou, and D.~Lischinski, ``Crowds by example,'' in
  \emph{Computer graphics forum}, vol.~26, no.~3.\hskip 1em plus 0.5em minus
  0.4em\relax Wiley Online Library, 2007, pp. 655--664.

\bibitem{kingma14-adam}
D.~P. Kingma and J.~Ba, ``Adam: A method for stochastic optimization,''
  \emph{arXiv preprint arXiv:1412.6980}, 2014.

\bibitem{yao19-group}
X.~Yao, J.~Zhang, and J.~Oh, ``Following social groups: Socially compliant
  autonomous navigation in dense crowds,'' \emph{arXiv preprint
  arXiv:1911.12063}, 2019.

\end{thebibliography}
\newpage
\end{document}